\documentclass[11pt]{article}

\usepackage[margin=1in]{geometry}

\usepackage[T1]{fontenc}
\usepackage{amsmath}
\usepackage{amsfonts}
\usepackage{graphicx}
\usepackage{xcolor}
\usepackage{hyperref}

\usepackage{cancel}
\usepackage{tikz}
\usepackage{comment}
\usepackage[normalem]{ulem}

\hypersetup{
    colorlinks=true,
    linkcolor=blue,
    citecolor=blue,
    urlcolor=blue
}

\title{Self-Organized Learning in Oscillatory Neural Networks with Memristive Signed Couplings}

\usepackage{authblk}

\title{Self-Organized Learning in Oscillatory Neural Networks with Memristive Signed Couplings}

\author[1]{Riley Acker}
\author[1,2]{Aman Desai}
\author[1]{Garrett Kenyon}
\author[3,*]{Frank Barrows}
\affil[1]{Computing and Artificial Intelligence (CAI) Division, 
Los Alamos National Laboratory, Los Alamos, NM 87545, USA}
\affil[2]{Center for Nonlinear Studies (CNLS), 
Los Alamos National Laboratory, Los Alamos, NM 87545, USA}
\affil[3]{Theoretical Division , 
Los Alamos National Laboratory, Los Alamos, NM 87545, USA}
\affil[*]{fbarrows@lanl.gov}

\date{}

\begin{document}

\maketitle

\begin{abstract}
Oscillatory neural networks (ONNs) have emerged as a promising neuromorphic architecture, leveraging coupled dynamical systems to perform computation and represent information through phase relationships. Their interactions can be designed to support intrinsic energy-minimizing dynamics, enabling tasks such as associative memory and optimization, and positioning them as a candidate architecture for continuous learning and inference. We present a neuromorphic primitive implemented using memristive edges with inhibitory couplings as a potential design for 
autonomous learning, and provide circuit simulation validation that the system is capable of denoising noisy inputs on an auto-associative task. While numerical Hopfield/Ising models routinely assume signed weights,
neuromorphic implementations of ONNs often fail to realize negative weights due to device and circuit constraints.  A practically implementable route to \emph{inhibitory (negative) weights} is particularly valuable: it expands the class of attractor structures accessible to oscillator networks beyond purely synchronous couplings, and supports phase-coded memories where anti-phase constraints are not merely transiently enforced during training but can persist autonomously after release. We provide circuit simulations and theoretical analyses demonstrating that signed effective weights are necessary for anti-phase attractors to persist autonomously.

\end{abstract}
\section{Introduction}

Neuromorphic computing is a promising direction for next-generation computing architectures, drawing on principles from neuroscience to design adaptive systems that more closely mirror information processing in the brain \cite{Indiveri_IEEE_2015,Kudithipudi2025}. Computing paradigms based on conventional learning algorithms, such as backpropagation, are often incompatible with neuromorphic hardware due to their reliance on non-local update rules; such algorithms are instead typically implemented on von Neumann architectures, where memory and computation are physically separated, requiring repeated, energetically costly data movement between storage and processing units.  
In contrast, biological neural networks embed memory directly in the weights of the synapses between neurons, enabling memory and computation to be co-located within the same physical substrate \cite{hebb_book,hopfield_networks}. Neuromorphic hardware can be configured to physically realize this co-location \cite{Indiveri_IEEE_2015}, enabling computation to arise from local dynamics governed by device physics. 

Such neuromorphic approaches are naturally described by dynamical systems, as their computation emerges from the evolution of state variables over time. While many leading neuromorphic approaches rely on spiking neural networks (SNNs) \cite{exploring_neuro_computing}, equally important are the rich collective dynamics that emerge at the neural population level \cite{population_doctrine}. Neural activity exhibits widespread oscillations and synchronization \cite{neuronal_oscillations,patterns_oscillatory_amplitudes}, and both experimental and theoretical studies suggest that phase \cite{theta_phase_procession,theta_rhythm,what_is_a_moment} and attractor dynamics \cite{hopfield_networks,attractor_dynamics_neocortex} play key roles in distributed memory and computation. We explore this abstraction beyond spiking, where computation and memory emerge through relaxation toward stable attractor states in an oscillating system.

In hardware instantiations, such dynamics require devices with activity-dependent state evolution. Memristors are candidate two-terminal devices whose conductance evolves as a function of the time-integral of applied voltage or current \cite{memristor_chua}. When configured in a network, these plastic weights can change over time and implicitly define an energy landscape for computation. In ONNs, this enables phase-dependent plasticity, an analog to spike-timing dependent plasticity (STDP), and co-locates both learning and inference within the intrinsic device physics. 

ONNs have long been studied as associative-memory systems in which information is stored in stable phase-locked relationships, extending the Hopfield memory paradigm from static fixed points to oscillatory attractors \cite{oscillatory_neurocomputers,Aoyagi,NISHIKAWA2004134}. Prior work on coupled-oscillator learning and synchronization shows that the sign of coupling is critical: positive interactions favor in-phase locking, whereas negative interactions stabilize anti-phase relationships \cite{PhysRevE.84.046202,PhysRevLett.106.054102}.


In analog ONN implementations, true negative resistive couplings are often difficult to realize under hardware constraints and use only positive couplings \cite{shamsi2021hardware}; we find that these negative weights are necessary for stable anti-phase relationships. 
In ONNs with weights implemented with memristive elements, phase locking minimizes the instantaneous voltage drop between oscillators, making it challenging to trigger effective phase-dependent weight updates as there is no forcing signal when oscillators are synchronized.

In this work, to overcome this, we couple Wien-bridge oscillators with a static resistive element placed in the non-inverting path, and a memristive element placed in the inverting path. This enables native Hebbian-like plasticity and for synapses to effectively take on both negative and positive weights. A key hardware contribution is a compact \emph{signed} coupling primitive for oscillator neural networks. We refer to each oscillator as a ``\textit{neuron},'' net positive connections as ``\textit{excitatory},'' and negative weights as ``\textit{inhibitory}''.

This oscillator-memristor platform realizes \emph{self-organized learning} as a coupled nonlinear dynamical system in hardware which is stabilized by mixed inhibitory and excitatory interactions: phases evolve continuously under a Kuramoto-type interaction while each synapse evolves continuously under a 
memristor state equation. Learning is not implemented by a digital weight update rule or an external clock; rather, it arises from the closed-loop coevolution of phase and conductivity, $(\phi,x)$, where phase relations generate local device currents that adapt conductances, and the adapted conductances reshape the attractor landscape that governs subsequent phase dynamics. This perspective, where training relates to the stability of coupled ODEs on a physical network, provides a natural theoretical language for both successful recall and failure modes when scaling pattern sets.

\section{System Overview}
\begin{figure}[htbp]
  \centering
  \includegraphics[width=0.5\linewidth]{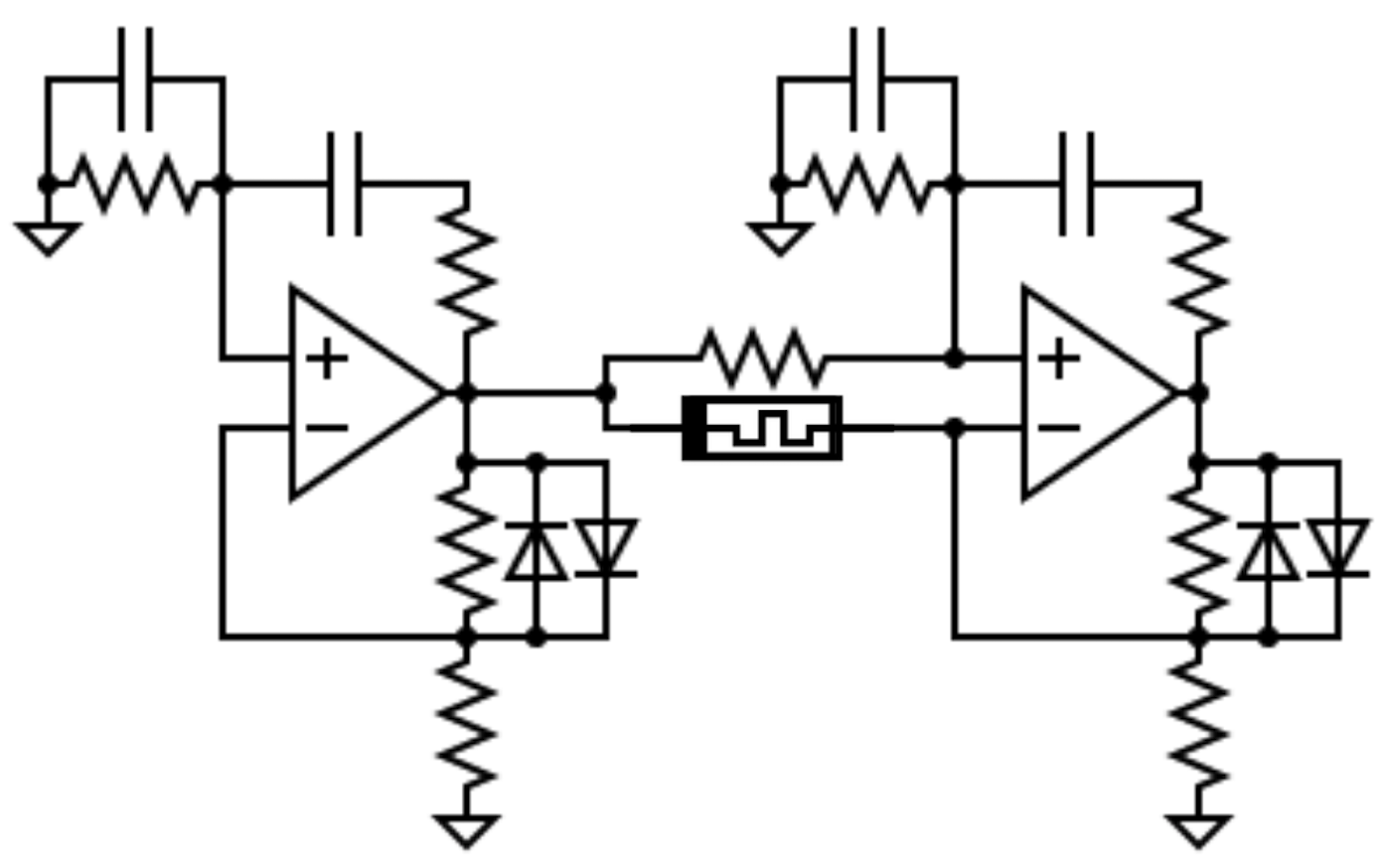}
  \caption{Circuit schematic of two Wien bridge oscillators coupled by resistors and memristors. Excitatory coupling  is implemented with resistors via non-inverting inputs, while adaptive inhibitory coupling is implemented with memristors via inverting inputs, enabling both positive and negative effective weights. Memristor conductance evolves with the voltage difference between oscillators, producing phase-dependent plasticity.}
  \label{fig:circuit}
\end{figure}

In these experiments, we utilize the Wien bridge oscillator, a classic analog circuit used in low-distortion signal generation applications. These oscillators are also easily prototyped in primitive circuit demonstrations and  approximate the Kuramoto model \cite{emergence_analysis_ks_wien_bridge}, a nonlinear dynamical system describing how a system of coupled oscillators synchronize through phase interactions according to their coupling strengths and natural frequencies \cite{kuramoto_model}. Wien bridge oscillators produce stable sinusoidal signals of a fixed amplitude with a frequency governed by their lead-lag RC network. Coupling oscillators through the non-inverting op-amp input synchronizes phase, while coupling through inverting input drives a $\pi$ phase offset between oscillators. These oscillators are connected via a memristor and a static resistor wired to the inverting and non-inverting inputs, respectively. As shown in Fig.~\ref{fig:circuit}, the resistor connecting the oscillators should have a resistance that is less than the high resistance state (HRS) of the memristor but higher than the low resistance state (LRS), allowing the couplings between oscillators to take on both positive and negative weights. In this particular setup, we use volatile memristors with reset thresholds that are never reached.

In ONNs, Hebbian learning can be implemented through phase-difference interactions, where coupling depends on the cosine of phase differences between oscillators \cite{oscillatory_neurocomputers}. In our primitive neuromorphic circuit demonstration, when the phase difference between out-of-phase oscillators (realized physically as a difference in voltage) exceeds the set threshold, the memristor is driven to a lower resistance state. As the instantaneous voltage difference is governed by their relative phase, the resulting weight updates follow an ``anti'' Hebbian rule, wherein oscillators that are sufficiently out of phase induce potentiation of the inhibitory coupling, reinforcing their phase separation. When the phase difference is small and the threshold is not exceeded, the memristor relaxes toward a higher resistance state, reducing the inhibitory coupling and biasing the interaction toward synchronization. 

\section{Circuit Simulations}
Circuit experiments were performed in LTspice, a SPICE (Simulation Program with Integrated Circuit Emphasis) simulation tool. We first construct a fully connected network of four Wien bridge oscillators, which act as neurons in a network. 

In the following simulations, the Wien bridge feedback network uses $10\,\mathrm{k\Omega}$ resistors and $1\,\mathrm{\mu F}$ capacitors to achieve a frequency of approximately $16\,\mathrm{Hz}$, given by $f = \frac{1}{2\pi \cdot (1\,\mathrm{\mu F}) \cdot (10\,\mathrm{k\Omega})}$. For physical realism, we use the TLV2461 op-amp macromodel provided by Texas Instruments.


The modified memristor model \cite{Amer:2017} used in these demonstrations implements a voltage threshold, state-dependent HfO$_2$ device in which conductance evolves according to an internal state variable governed by a first-order nonlinear ODE.

\subsection{Memristive Hebbian Learning}
To verify that these primitive circuits with memristive elements are capable of Hebbian-like learning of weights for an oscillatory implementation of a Hopfield network \cite{hopfield_networks}, we first perform circuit simulations of four oscillators in a fully connected network. The op-amps are powered by ±1.5V supply rails, and antiparallel diodes are used as automatic gain control on the amplitude of oscillators. Fully connected networks are connected by symmetric, reciprocal edges. Each oscillator pair is connected by two reciprocal edges, each carrying its own memristor. Because the magnitude of voltage-difference across each memristor depends on relative phase, $|\phi_i - \phi_j|$, which is effectively the same in both directions, the two memristors in each pair evolve to the same conductance state, yielding symmetric weights $w_{ij} \approx w_{ji}$.

We first demonstrated training patterns in a small oscillator network, where a pattern is a bit string we intend to encode, represented as a vector of four oscillator phases. Patterns and phases can be interpreted as logical binary data corresponding to phases of $0$ and $\pi$ relative to the rotating frame. Idealized voltage sources drive this 4 bit sinusoidal pattern into the network, synchronizing or anti-synchronizing all oscillators. The network is subjected to external driving for $5\,s$ according to the target pattern to be learned, after which the forcing is removed, all oscillators are briefly grounded, i.e., phase reset, and the system freely evolves under the learned synaptic weights for another $5\,s$.

\begin{figure}[htbp]
  \centering
  \includegraphics[width=0.7\linewidth]{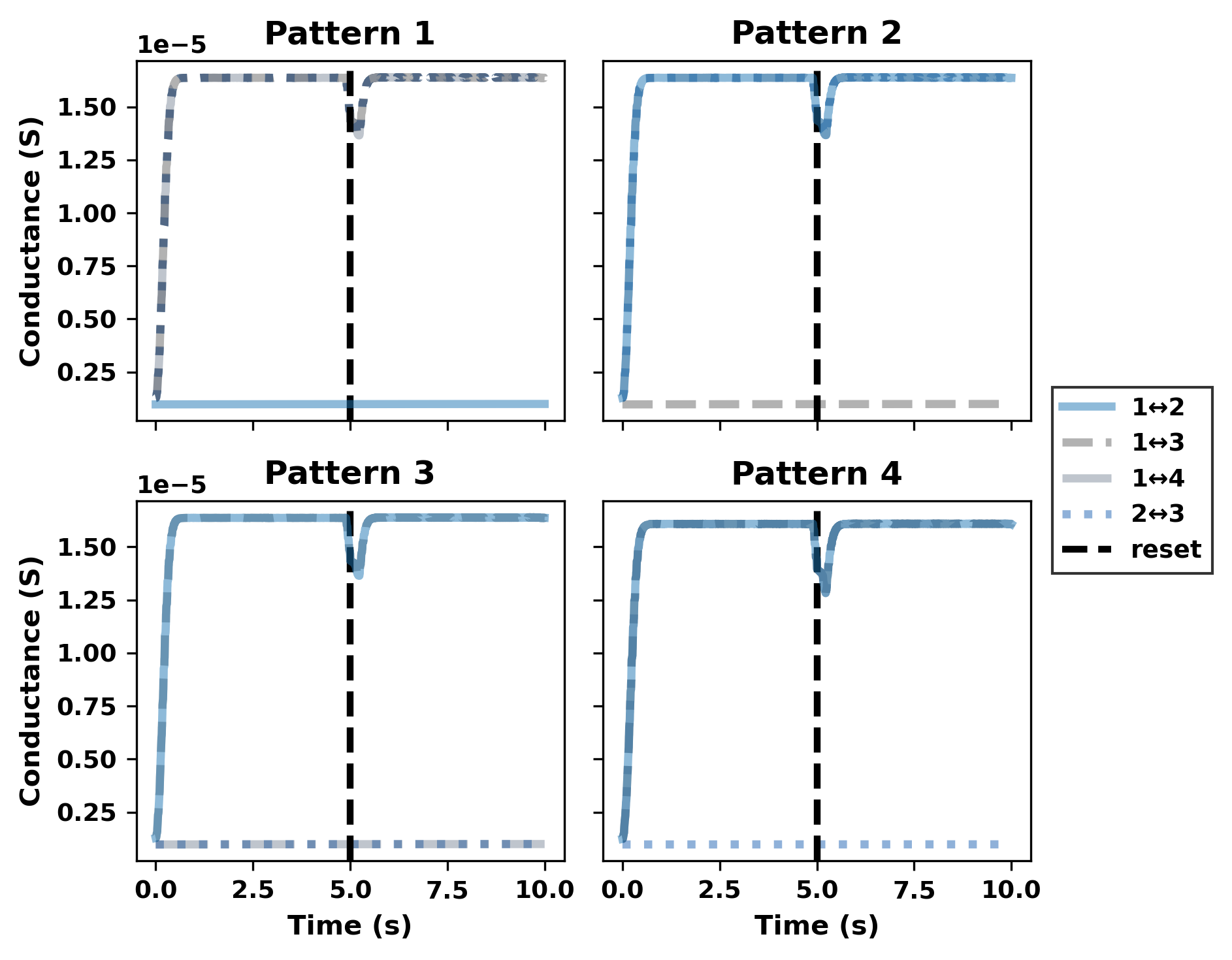}

  \caption{Demonstration of pattern learning and recall in a fully connected four oscillator network. Four representative phase patterns $\mathbf{[0,0,\pi,\pi]}$, $\mathbf{[0,\pi,0,\pi]}$, $\mathbf{[\pi,0,0,\pi]}$, and $\mathbf{[\pi,0,0,0]}$
  are encoded by externally driving oscillator phases for the first \textbf{$5\,s$}, during which pairwise memristor conductances evolve to store coupling weights. External clamping is removed and network evolves under the learned weights, recalling stored phase relationships.}
  \label{fig:Learning4Patterns}
\end{figure}

Eight distinct patterns are realizable in this four-oscillator demonstration, shown in Fig.~\ref{fig:Learning4Patterns}. In pattern 1, $[0,0,\pi,\pi]$, the first two oscillators are externally driven to be the same phase, while the remaining two oscillators are clamped at $\pi$ phase difference relative to the first two. Synaptic weights, instantiated in the memristor conductance, are learned continuously during the forcing period. After the reset at $5\,s$, the network returns to the previously learned configuration, consistent with a learned attractor.

\subsection{Necessity of Inhibitory Couplings}
With Hebbian-like learning demonstrated in an ONN with inhibitory weights, we next compare the retrieval performance shown in Section 3.1 to a corresponding excitatory-only network, in which plastic weight updates cannot be realized due to the absence of the memristive inhibitory pathways, and weights are instead instantiated through static couplings.
In the non-inhibitory network, only non-inverting (positive) pathways are used, and resistors are chosen to match the effective resistance range of the memristors. In this excitatory-only configuration, positive couplings derived from the ideal Hebbian couplings are mapped to the lowest realizable resistance, while negative couplings are set to be the highest resistance. 

\begin{figure}[htbp]
  \centering
  \includegraphics[width=0.8\linewidth]{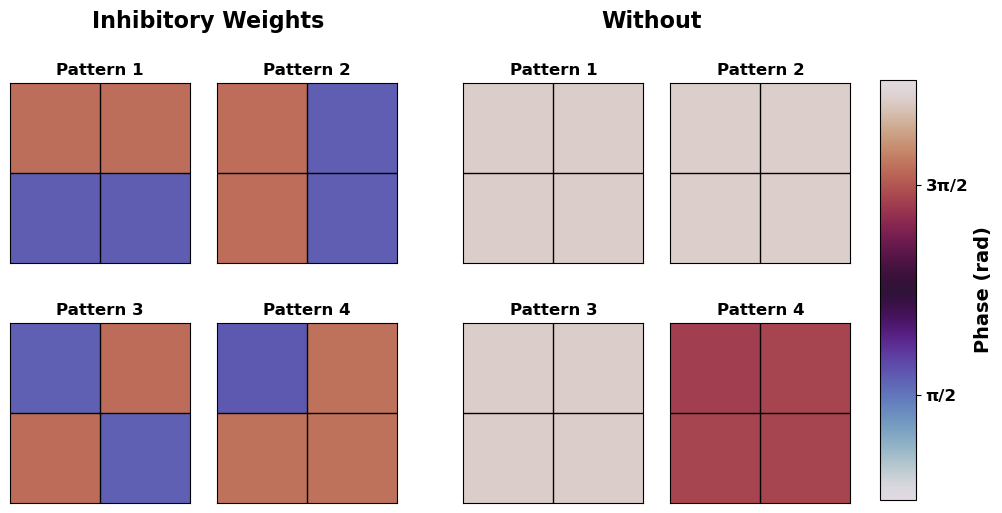}
  \caption{Phases of the oscillators at the end of the recall period. The network is simulated with and without inhibitory weights. For each of the four patterns,  phases are imposed during training and then allowed to evolve freely, and phases at the last timestep in the simulation are displayed in a grid, ordered left-to-right.}
  \label{fig:recall4patterns}
\end{figure}

We encode all four patterns from Fig.~\ref{fig:Learning4Patterns} in two networks: one with inhibitory weights and one without. For the network with inhibitory edges, inhibitory and excitatory weights at the final timestep $(10\,s)$ are shown in Figure 2. For the network without inhibitory couplings, edge weights are computed using the standard Hopfield outer-product rule for each pattern; these weights are then mapped to resistances achievable by the memristor used in the network with inhibitory edges, where a positive weight corresponds to the resistance at the memristor LRS, and a negative weight to the achievable HRS. Phases plotted after $5\,s$ of recall are shown in Fig.~\ref{fig:recall4patterns}. Networks without inhibitory weights fail to recall any pattern containing anti-phase ($\pi$) relationships between oscillators and collapse into a fully synchronized steady-state, except for a slight phase deviation in the first oscillator for Pattern 4. 

\begin{figure}[htbp]
  \centering
  \includegraphics[width=0.8\linewidth]{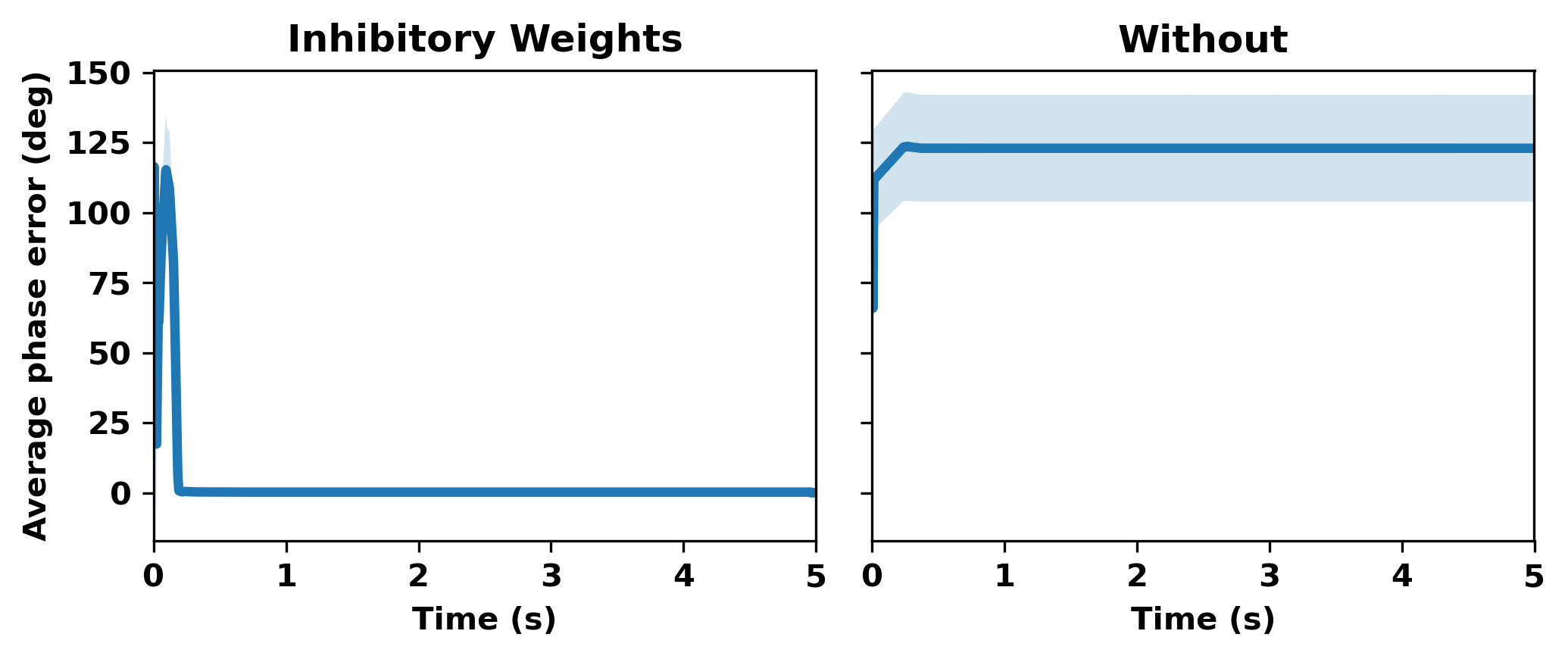}
  \caption{Mean pairwise phase error during recall averaged across the four trained patterns and compared for networks with and without inhibitory weights. Shaded regions denote standard deviation.}
  \label{fig:relativephase_recall}
\end{figure}

To quantify retrieval performance in networks with and without inhibitory couplings, we compute the mean phase error between the network state and the target pattern during the recall phase. For each time point, pairwise phase differences between all oscillator pairs are calculated and compared to the corresponding target phase differences. The error for each pair is defined as the wrapped angular difference in degrees, with values within $[-180^\circ, 180^\circ]$. The absolute error is then averaged across all oscillator pairs to obtain a single phase error value at each time point, and is normalized to a percentage by dividing by the maximum possible phase deviation ($180^\circ$), yielding a percent phase error metric, shown in Fig~\ref{fig:relativephase_recall}. Under this formulation, $0\%$ corresponds to perfect phase alignment, while $100\%$ represents maximal disagreement. At the last timestep in the circuit simulation, networks with inhibitory weights achieve near-zero average error ($\approx 3.0 \times 10^{-9}\%$), whereas networks without inhibitory weights exhibit substantially higher error ($\approx 62\%$), corresponding to a average phase error of $0^\circ$ and $111^\circ$, respectively, and demonstrating the necessity of inhibitory couplings in stabilizing correct phase relationships.

In the excitatory-only network, all four patterns converge to approximately the same fully synchronized state regardless of the trained pattern. This is expected: with only positive couplings, every edge favors in-phase locking, so the system relaxes to a trivial synchronized configuration in which all oscillators share the same phase. Anti-phase relationships, which require that some edges actively resist synchronization, cannot be sustained without negative effective couplings. The theoretical basis for this observation is developed in Section~\ref{sec:theory}.

\subsection{Autonomous Learning of Binary Numbers} 
Here, we extend the memristive-Hebbian network to a small learning task, where oscillators arranged in a fully connected $3 \times 3$ grid are trained to store and recall binary patterns corresponding to the digits ``0'' and ``1'', using the same circuit architecture and general methods previously described.

\begin{figure}[htbp]
  \centering
  \includegraphics[width=0.6\linewidth]{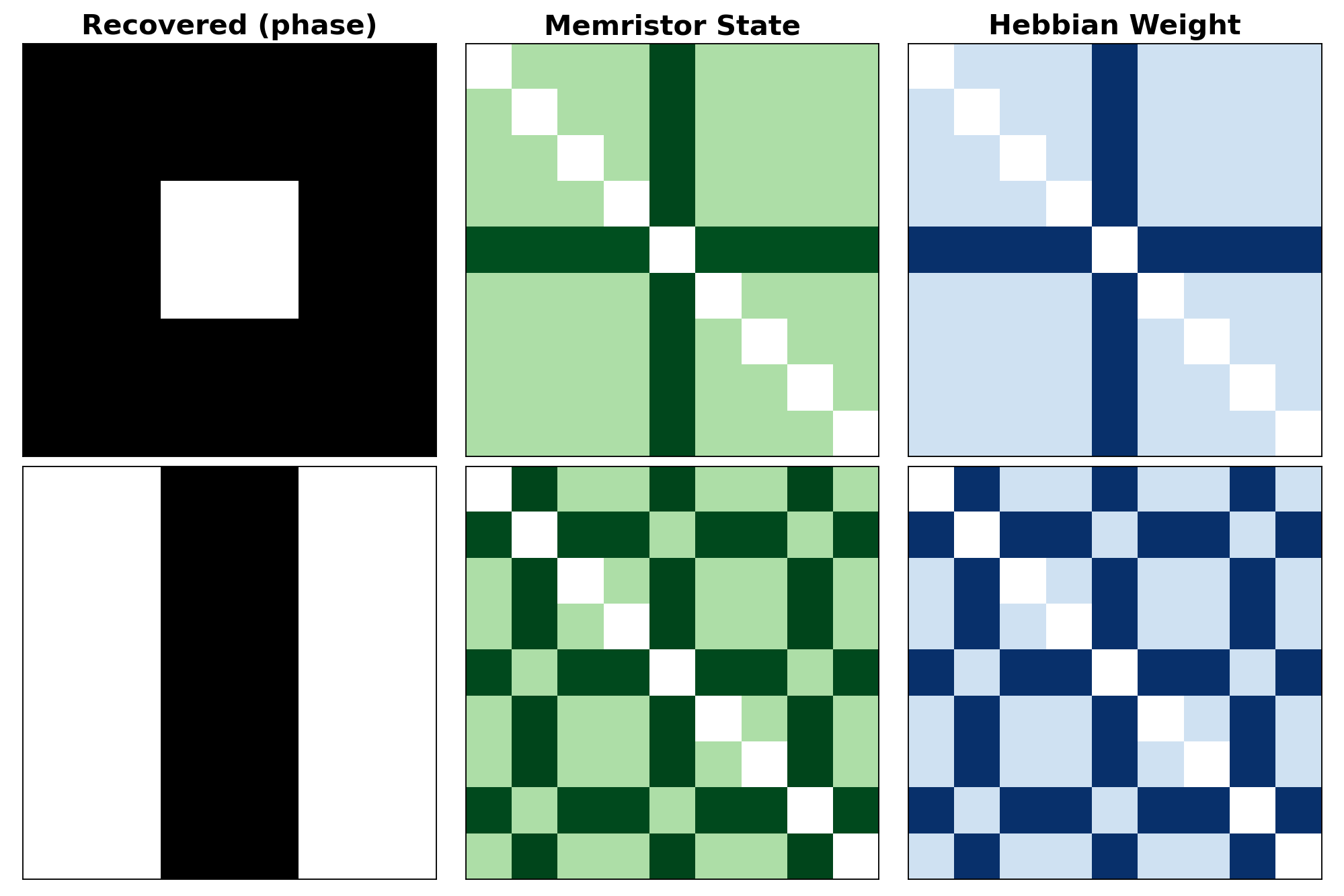}
  \caption{Memristor-based Hebbian learning for the digits 0 and 1 in a 3×3 fully connected oscillatory network. First column: recovered binary phase pattern at the end of recall for the 9 oscillators, displayed in a 3×3 grid. Second column: final memristor states on inhibitory connections, lighter green corresponds to the HRS and darker green to the LRS. Third column: ideal Hebbian weight matrix, where darker blue denotes negative weights and lighter blue denotes positive weights.}
  \label{fig:MemristorLearning}
\end{figure}

Oscillator phases were extracted using zero-crossing interpolation, wherein sine waves are interpolated between times at which voltages cross zero, and final network states were binarized based on relative phase differences. Learned synaptic weights were inferred from final memristor states and compared to ideal Hebbian weights computed as the outer product of the target patterns (Fig.~\ref{fig:MemristorLearning}). Recall accuracy at the final timestep in the simulation was quantified by comparing measured pairwise phase differences to the ideal $0^\circ$ and $180^\circ$ relationships using the mean pairwise absolute phase error, yielding an error of $0.7^\circ$ (0.40\%) for ``0'' and $1.2^\circ$ (0.66\%) for ``1''. We compare the theoretical ideal Hebbian weights to the measured memristor states and find that they are in direct agreement, demonstrating that the memristive synapses are capable of Hebbian-like learning.

\begin{figure}[!htbp]
  \centering
  \includegraphics[width=0.6\linewidth]{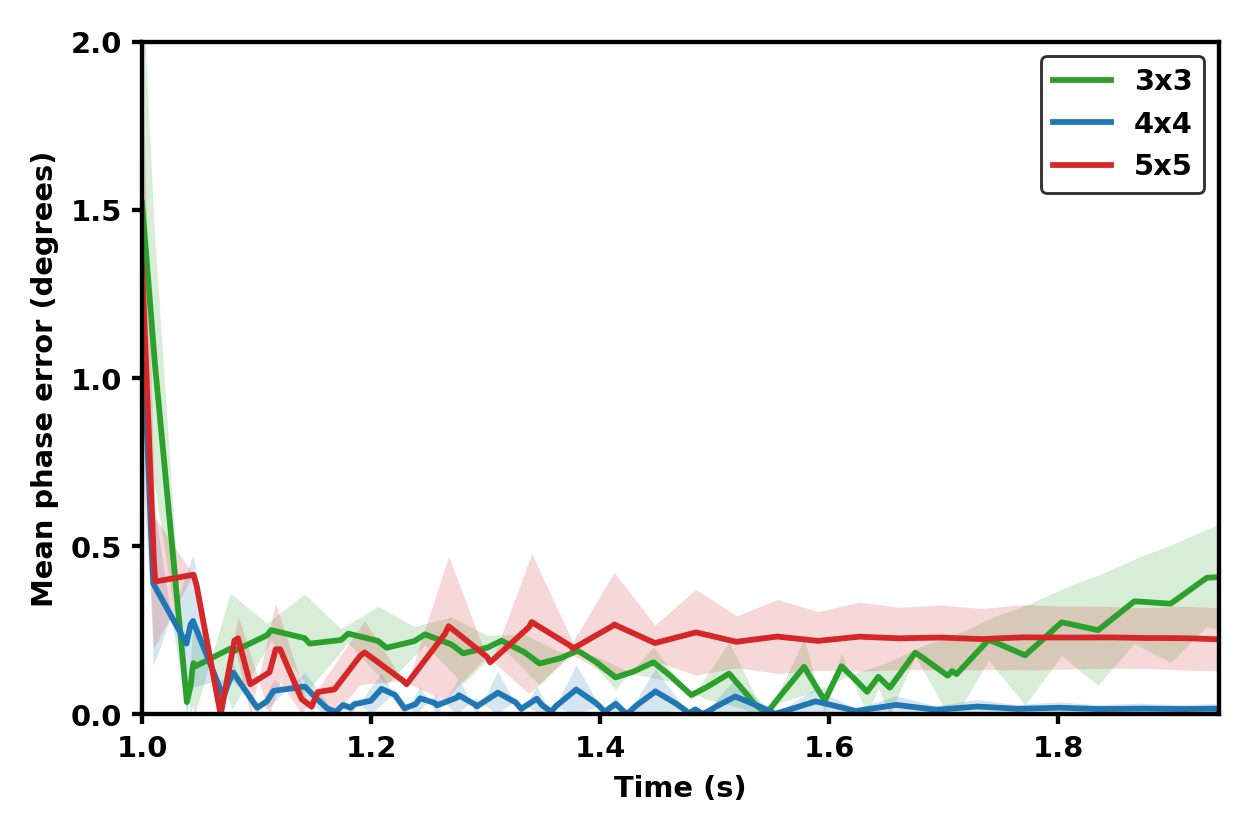}
  \caption{Mean phase error over a 1 $s$ inference period following memristive Hebbian learning. Errors for the digit patterns 0 and 1 are averaged and shown for fully connected oscillatory networks of size 3×3, 4×4, and 5×5. Shaded regions denote standard deviation.}
  \label{fig:345_errors}
\end{figure}

We extend these $3 \times 3$ demonstrations to larger $4 \times 4$ and $5 \times 5$ networks, with average error for both digits at inference shown in Fig.~\ref{fig:345_errors}. The $4 \times 4$ network achieved mean pairwise absolute phase errors of $0.07^\circ$ for ``0'' and $0.04^\circ$ for “1”, corresponding to an average relative error of 0.03\% throughout the 1 $s$ inference period. The $5 \times 5$ network exhibited errors of $0.3^\circ$  for “1” and $0.1^\circ$ for “0”, yielding an average error of 0.1\% across both digit patterns. We find that error settles when ran for greater than the displayed 1 $s$ inference window.

\subsection{Denoising the Digits Dataset}
Here we demonstrate auto-associative memory in a circuit with both positive and negative coupling implemented with fixed resistances. 
A 64 neuron oscillatory network was implemented to perform an $8\times 8$ pixel reconstruction of handwritten digits dataset. In this configuration, the network encodes six unique “memories,” and recall performance is evaluated based on whether the phase configuration converges to the nearest stored pattern under noisy initialization. Idealized op-amp models were used in SPICE, and resistive elements were employed for all synaptic connections.

Prototype images were created by binarizing and flattening the average of images in each of the first six digit classes into a 64-element vector. These stored patterns were combined into a symmetric weight matrix, which was ternarized within \{-1,0,1\} and then sparsified by masking connections beyond a fixed Euclidean radius on the 2D grid, yielding a more locally connected network. 
To map these idealized Hebbian weights to circuit parameters, each synaptic connection was implemented through parallel resistive pathways to the non-inverting and inverting inputs of the downstream oscillator. A fixed resistance was used on the non-inverting pathway, while the inverting pathway resistance was selected according to the ternary weight value, with positive and negative weights realized as the ratio of the resistance of the inverting path compared to the non-inverting. 

At the start of each simulation, a corrupted cue was generated by initializing the network with a stored digit, where binarized pixel values were mapped to oscillator phases of $0$ or $\pi$ radians relative to a rotating reference frame, and noise was introduced by randomizing the phases of 15$\%$ of the oscillators uniformly over the unit circle. The corrupted cue is momentarily applied to the network, after which external forcing is removed and the circuit evolves freely for 0.6$s$. Oscillator phases were embedded as points on the unit circle, and KMeans clustering was used to identify the two dominant phase groups for plotting. The oscillator network reconstructs all of the six cued digits an average error of $2\%$ across all digits.

\begin{figure}[!htbp]
  \centering
  \includegraphics[width=0.8\linewidth]{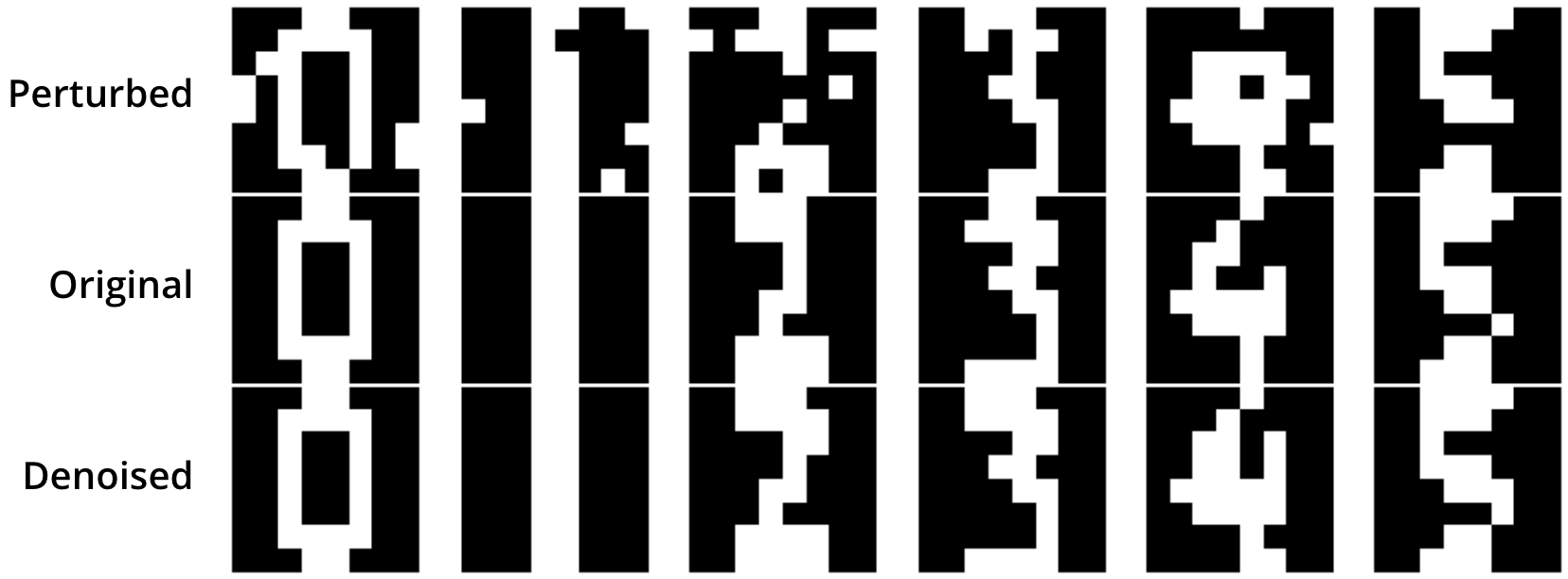}
  \caption{Pattern reconstruction in a 64-neuron SPICE simulation on a handwritten digits autoassociative memory task. The network stores six $8 \times 8$ handwritten digit prototypes as phase-coded memories in a resistive weight matrix. In each example, the first row shows the noisy cue, the second row shows the corresponding stored prototype, and the third row shows the pattern recovered after the network evolved under learned weights. Oscillator phases encode binary pixel states as $0$ or $\pi$, and reconstruction is evaluated by whether the final phase configuration converges to the phase-representation of the appropriate digit.}
\end{figure}

\section{Theoretical underpinnings: Coupled oscillator-memristor dynamics and stability}
\label{sec:theory}
This section formulates the oscillator-memristor network as a coupled (autonomous) ODE and gives stability conditions for
phase-locked attractors in the  coupled phase-conductance $(\phi,x)$ picture.  The key point is that polarity-sensitive memristor dynamics is driven by
a periodic edge current, so stability is naturally characterized by Floquet exponents.

\subsection{Phase reduction with state-dependent couplings}
We model $N$ oscillators with phases $\phi_i(t)\in\mathbb{R}$ coupled through symmetric effective weights
$K_{ij}(x_{ij})=K_{ji}(x_{ji})$ that depend on memristor states $x_{ij}(t)\in[0,1]$ on edges $(i,j)\in E$:
\begin{equation}
\dot{\phi}_i \;=\; \omega_i \;-\; \sum_{j\neq i} K_{ij}(x_{ij})\,\sin(\phi_i-\phi_j).
\label{eq:kuramoto_mem}
\end{equation}
In the homogeneous rotating frame one takes $\omega_i\equiv 0$, and for \emph{frozen} $x$ the phase dynamics are equivalent to a gradient flow of an energy function
\begin{equation}
F(\phi\,|\,x)\;=\;-{1 \over 2}\sum_{i<j}K_{ij}(x_{ij})\,\cos(\phi_i-\phi_j).
\label{eq:energy_conditional}
\end{equation}
Local stability of a phase-locked state for fixed $x$ is therefore controlled by the curvature (Hessian) of $F(\cdot|x)$; when synapses co-evolve, the appropriate generalization is Floquet stability of the resulting time-periodic system, developed below.

\subsection{Current-driven memristor model and autonomous lifting}
We take a standard linear-in-memory resistance model
\begin{equation}
R(x)\;=\;R_{\mathrm{off}}(1-x)+R_{\mathrm{on}}x,
\qquad
G(x)\;=\;R(x)^{-1},
\label{eq:R_of_x}
\end{equation}
so $G(x)$ is nonlinear in $x$.  The internal state, i.e., memory parameter $x$, is driven by the \emph{signed} device current,
\begin{equation}
\dot{x}_{ij} \;=\; f\!\big(i_{ij}(t),x_{ij}\big),
\label{eq:mem_current_general}
\end{equation}
where $f$ encodes polarity dependence (set/reset asymmetry, thresholds/windowing) and volatility/leak is included as part of $f$. Here we set $R(x=1)=R_\text{LRS}$ and $R(x=0)=R_\text{HRS}$.

To couple \eqref{eq:mem_current_general} to the phase model we use the standard phase-reduction assumption that each oscillator runs on a
stable limit cycle with approximately constant amplitude $A$ at carrier frequency $\Omega$,
\begin{equation}
V_i(t)\approx A\cos(\Omega t+\phi_i(t)).
\label{eq:Vi}
\end{equation}
For a two-terminal memristor oriented from node $i$ to node $j$, the edge current is
\begin{equation}
i_{ij}(t)
=
\frac{V_i(t)-V_j(t)}{R(x_{ij})}
=
2A\,G(x_{ij})\,\sin\!\Big(\frac{\phi_i-\phi_j}{2}\Big)\,
\sin\!\Big(\Omega t+\frac{\phi_i+\phi_j}{2}\Big).
\label{eq:iij_from_phase}
\end{equation}
Because $i_{ij}(t)$ is explicitly periodic, the natural autonomous formulation augments the state by the carrier phase
\begin{equation}
\theta:=\Omega t,
\qquad \dot\theta=\Omega,
\label{eq:theta}
\end{equation}
so that the coupled oscillator-memristor network is an autonomous ODE in $(\phi,x,\theta)$.

\subsection{Signed effective weights from excitatory-inhibitory pathways}
The circuit motif used here implements a fixed excitatory pathway in parallel with an adaptive inhibitory pathway.  At the phase level
we write
\begin{equation}
K_{ij}(x)\;=\;K^{(+)}_{ij}\;-\;K^{(-)}_{ij}(x),
\qquad
\frac{d}{dx}K^{(-)}_{ij}(x)\ge 0,
\label{eq:signed_K}
\end{equation}
with $K^{(-)}_{ij}(x=0)<K^{(+)}_{ij}$ so that increasing inhibitory strength can drive $K_{ij}(x)$ through zero.  A concrete instance is
$K^{(-)}_{ij}(x)=\gamma_{ij}G(x)$ with $\gamma_{ij}>0$.

\subsection{Two-oscillator reduction and Floquet exponents}
The essential coupled-picture stability mechanism is already visible for a single memristive edge connecting two oscillators.
In the homogeneous rotating frame ($\omega_1=\omega_2$), define $\Delta:=\phi_1-\phi_2$ and write the effective coupling as $K(x)$.
Then the phase difference obeys
\begin{equation}
\dot\Delta = -2K(x)\sin\Delta.
\label{eq:Delta_pair}
\end{equation}
The memristor sees the periodic current \eqref{eq:iij_from_phase}; using $\theta$ from \eqref{eq:theta} we may write
\begin{equation}
\dot x = f\!\left(2A\,G(x)\sin(\Delta/2)\sin(\theta+\sigma),x\right),
\label{eq:x_pair_forced}
\end{equation}
where $\sigma=(\phi_1+\phi_2)/2$ is constant in the symmetric two-oscillator reduction and can be absorbed into $\theta$. 
Thus the coupled pair dynamics is a $3$-dimensional autonomous system in $(\Delta,x,\theta)$.

The binary phase-locked manifolds $\Delta\equiv 0$ and $\Delta\equiv \pi$ are invariant for \eqref{eq:Delta_pair}.
On these manifolds, \eqref{eq:x_pair_forced} reduces to a scalar forced system $(x,\theta)$ with period $T=2\pi/\Omega$.

\paragraph{In-phase manifold.}
On $\Delta\equiv 0$ the edge current vanishes ($i_{12}(t)\equiv 0$), so $x$ follows the unforced dynamics $\dot x=f(0,x)$.
Assume it has an attracting equilibrium $x_0$ (set by device relaxation).  Perturbing $\Delta=\varepsilon$ gives
$\dot\varepsilon\approx -2K(x(t))\varepsilon$, hence the in-phase relation is attracting provided
\begin{equation}
\limsup_{t\to\infty}\frac{1}{t}\int_0^t K(x(s))\,ds \;>\; 0 .
\label{eq:lambda0}
\end{equation}

\paragraph{Anti-phase manifold.}
On $\Delta\equiv \pi$ the current is a full-amplitude sinusoid,
\[
i_{12}(t)=\frac{2A\,\sin\theta}{R(x)}.
\]
Assume the forced $(x,\theta)$ subsystem on $\Delta=\pi$ admits an attracting $T$-periodic solution $x_\pi(t)$ such that $x_\pi(t)=x_\pi(t+T)$. Physically this is a stable hysteretic steady state under sinusoidal drive.   
This is a valid assumption as under periodic drive we define a Poincar\'e  map, $\mathcal{P}$, wherein for a chosen pattern $\mu$ over a dwell interval $\tau$, 
$x\mapsto \mathcal{P}^{(\mu)}_{\tau}(x)$. We have a guarantee there exists at least one fixed point of the Poincar\'e map as the resistance is bounded, $x\in[0,1]$ and $\mathcal{P}$ is a continuous map for a drive that does not saturate $x$, corresponding to at least one periodic orbit.  The attractor orbit condition can be written as $|\mathcal{P}^\prime(x_\pi(0))|<1$, a sufficient condition is the dynamics are contractive such that 
\begin{equation}
   \frac{d}{dx}f\big(i_{ij}(t),x_{ij}\big) < 0 ,
\end{equation}
   a condition satisfied for volatile memristors.

Because the memristor state equation is driven by a $T$-periodic current, its linearized stability is governed by Floquet theory, the natural extension of eigenvalue analysis to linear ODEs with periodic coefficients. 
Floquet theory quantifies how perturbations evolve over one period using the Floquet exponent: of particular use to us will be the fact that when the system's Floquet exponent has a negative real part, perturbations decay and the system stabilizes.
Now perturbing the anti-phase condition, $\Delta=\pi+\varepsilon$ gives $\dot\varepsilon\approx 2K(x_\pi(t))\varepsilon$.  The corresponding Floquet exponent for $\varepsilon$ is
\begin{equation}
\lambda_\pi \;=\; \frac{2}{T}\int_0^T K\!\big(x_\pi(t)\big)\,dt,
\label{eq:lambda_pi}
\end{equation}
so the anti-phase relation is attracting iff $\lambda_\pi<0$, equivalently
\begin{equation}
\frac{1}{T}\int_0^T K\!\big(x_\pi(t)\big)\,dt \;<\; 0.
\label{eq:pi_stable_condition}
\end{equation}
For unsigned couplings ($K(x)\ge 0$) this cannot hold, so $\Delta=\pi$ is not an autonomous attractor after release.  For signed effective weights \eqref{eq:signed_K}, anti-phase training can drive $x_\pi(t)$ into a regime where the cycle-average of $K$ is negative, leading to attractive dynamics and  stabilizing $\Delta=\pi$.

\subsection{Network extension: time-periodic curvature and stability of phase-coded patterns}
For a network of $N$-oscillators, consider a candidate phase-locked pattern $\vec{\phi}^*$ (e.g.\ a $0/\pi$-encoded memory encoded as an $N$-vector) and the corresponding edge
differences $\vec{\Delta}^*_{ij}=\vec{\phi}_i^*-\vec{\phi}_j^*$.  Along the coupled dynamics, each edge state $x_{ij}(t)$ evolves under its own
periodic drive determined by \eqref{eq:iij_from_phase} with $\Delta_{ij}\approx \Delta^*_{ij}$.
When $\Delta^*_{ij}\in\{0,\pi\}$, each edge is driven either by zero current (in-phase) or by a sinusoid of maximal amplitude
(anti-phase), so $x_{ij}(t)$ converges to a constant or a $T$-periodic orbit, respectively. 

Linearizing the phase dynamics about $\vec{\phi}^*$, $\vec{\phi}=\vec{\phi}^*+\vec{u}$,  we can write the phase dynamics as
\begin{equation}
    \dot{\phi}_i =  -\sum_j K_{ij}(x_{ij}(t))\big( \sin(\Delta^*_{ij}) +\cos(\Delta^*_{ij})(u_i-u_j)+ O(\|u\|^2) \big) .
\end{equation}
For binary patterns, $\sin(\Delta^*_{ij})=0$ so we have the form 
\[
\dot{u}_i=-\sum_j K_{ij}(x_{ij}(t))\cos(\Delta^*_{ij})(u_i-u_j).
\]
Then with the synapses evolving on their, possibly time dependent, steady states yields a linear time-periodic system 
\begin{equation}
\dot{\vec{u}} \;=\; -\,L(t)\,\vec{u},
\;\;
L(t)=B\,\mathrm{diag}\big(\vec{w}(t)\big)\,B^\top,
\;\;
w_e(t)=K_e(x_e(t))\cos(\Delta_e^*),
\label{eq:lin_network_periodic}
\end{equation}
where terms with subscript $_e$ correspond edge $e$ and $B$ is an incidence matrix \cite{discrete_calculus}.  This decoupling treats each $x_{ij}(t)$ as following its unperturbed steady-state orbit, neglecting the feedback $u \to \delta i \to \delta x \to \delta K$; this is valid when the memristor relaxation timescale is fast relative to the phase dynamics, so that the evolution of synapse perturbations can be modeled in terms of the current phase state.

Stability of $\vec{\phi}^*$ is governed by Floquet theory
for \eqref{eq:lin_network_periodic}.
To state the stability condition, decompose the perturbation as
$\vec{u}=\bar{u}\,\mathbf{1}+\vec{u}_\perp$, where
$\mathbf{1}=(1,\dots,1)^\top$ spans the global rotation mode which is always neutrally stable since $L(t)\mathbf{1}=0$ for any graph Laplacian, and $\vec{u}_\perp$ captures relative phase perturbations between oscillators.

A sufficient condition for stability, i.e., contraction of $\vec{u}_\perp$, is a uniform curvature bound: if there exists
$w_{\min}>0$ such that
\begin{equation}
w_e(t)\ge w_{\min}
\quad\text{for all edges }e\text{ and all }t,
\label{eq:uniform_curvature}
\end{equation}
then $L(t)$ is positive semidefinite with one-dimensional null
space $\mathrm{span}(\mathbf{1})$ for all $t$, and
$\vec{u}_\perp(t)$ decays exponentially at a rate controlled by
$w_{\min}$ and the graph connectivity.

For binary phase codes $\Delta^*_e\in\{0,\pi\}$ with
$\cos(\Delta^*_e)=\pm 1$, the curvature weights reduce to
\begin{align}
\Delta^*_e=0   &\;\Longrightarrow\;
  w_e=+K_e(x_e(t)),
  &&\text{stabilizing iff } K_e>0,
\label{eq:w_inphase}\\
\Delta^*_e=\pi &\;\Longrightarrow\;
  w_e=-K_e(x_e(t)),
  &&\text{stabilizing iff } K_e<0.
\label{eq:w_antiphase}
\end{align}
In an unsigned network ($K_e\ge 0$ everywhere), every anti-phase edge contributes $w_e\le 0$, rendering $L(t)$ indefinite and the stored pattern is unstable. No amount of graph connectivity can compensate, as positive semidefiniteness requires $w_e>0$ on every edge. This is the precise sense in which signed effective weights are
necessary for pattern stability: negative couplings on anti-phase edges are required for those edges to contribute positive curvature rather than destabilize the stored pattern.
 Successful learning therefore corresponds to driving each anti-phase edge into the regime $K_e<0$.

\paragraph{Floquet stability and time-periodic curvature.}
When the pattern contains anti-phase constraints, the curvature weights $w_e(t)$ are $T$-periodic and the appropriate stability notion is Floquet stability of \eqref{eq:lin_network_periodic}. 
The Floquet exponents of $\dot{\vec{u}}=-L(t)\vec{u}$ are determined by the monodromy matrix $\Phi(T)$, where
$\dot\Phi=-L(t)\Phi$ with $\Phi(0)=I$.  If $L(t_1)$ and $L(t_2)$ commuted for all $t_1,t_2$, one would have $\Phi(T)=\exp(-\int_0^T L(s)\,ds)$ and the Floquet exponents  would be the eigenvalues of the time-averaged Laplacian. 
Generically this fails, since different edges evolve on distinct periodic orbits, so an exact integral formula is not available.

One can nevertheless bound the contraction rate.  At each instant, $L(t)$ retains the weighted-Laplacian quadratic form $\vec{u}^\top L(t)\,\vec{u}=\sum_e w_e(t)(u_i-u_j)^2$, so when the uniform bound \eqref{eq:uniform_curvature} holds, $L(t)$ is positive semidefinite on $\mathbf{1}^\perp$ for all $t$.  Let $\lambda_2(t)$ denote the smallest nonzero eigenvalue of $L(t)$, the instantaneous connectivity.  Then
\begin{equation}
|\vec{u}_\perp(t)|
\;\le\;
|\vec{u}_\perp(0)|\,
\exp\!\Bigl(-\int_0^t \lambda_2(s)\,ds\Bigr),
\label{eq:floquet_bound}
\end{equation}
so the Floquet exponent governing pattern stability satisfies
\begin{equation}
\lambda_{\mathrm{net}}
\;\le\;
-\,\frac{1}{T}\int_0^T \lambda_2(s)\,ds.
\label{eq:floquet_network}
\end{equation}
This time-averaged algebraic connectivity plays the role of the scalar Floquet exponent \eqref{eq:lambda_pi} at the network level: it integrates the instantaneous graph curvature over one carrier period, and its sign determines whether the stored pattern is a stable attractor. 

\subsection{Implications for autonomous learning} 
The coupled system defined by \eqref{eq:kuramoto_mem}, \eqref{eq:mem_current_general}, and \eqref{eq:iij_from_phase} is fully autonomous and local: each synapse $x_{ij}$ evolves solely through the current set by its two terminal voltages, and the resulting conductance changes reshape the phase dynamics in return.   Learning requires no external weight-update rule or global clock. The Floquet stability conditions \eqref{eq:w_inphase}-\eqref{eq:w_antiphase} and the network contraction bound \eqref{eq:floquet_network} make precise which learned states support stable recall: in unsigned networks, 
anti-phase constraints cannot persist as attractors after the training signal is removed, whereas the excitatory-inhibitory motif permits $K_e$ to cross zero and renders those constraints self-sustaining.

\section{Discussion and Conclusions}
In this work, we demonstrate that ONNs with memristive inhibitory couplings can realize autonomous learning and inference, all within a single dynamical system. With continuous weight updates realized through phase-dependent plasticity in the inhibitory path through an excitatory-inhibitory balance, the proposed architecture  
allows for both positive and negative synaptic weights. This overcomes a key limitation in many hardware-realizable ONN implementations that are restricted to purely excitatory interactions. This expanded weight space allows the network to form stable phase-coded attractors supporting both in-phase and anti-phase relationships.

Circuit simulations verify that the memristor states evolve according to phase-dependent, local interactions, producing Hebbian-like weight updates that arise intrinsically from the system dynamics and device physics, as opposed to an externally imposed learning rule. Small scale networks demonstrate successful learning, with learned phase relationships being continuously reinforced as a stable attractor under the free system dynamics. In accordance with both our theoretical analyses and prior works, ablation results show that negative weights (inhibitory synapses) are essential \cite{negative_weights}: without them, the system cannot realize stable anti-phase relationships and can collapse into a fully synchronized state \cite{PhysRevE.84.046202,PhysRevLett.106.054102}, analogous to a ferromagnetic configuration with a trivial energy minimum incapable of learning multiple stable states.

Due to the computational cost of simulating large coupled networks with evolving memristive elements, learning is demonstrated only in the smaller networks, while the larger $8\times 8$ example uses fixed resistive couplings and idealized components. The present results should therefore be understood as a proof of concept for the signed-coupling mechanism and its ability to support attractor dynamics. We find that the associative memory properties of our platform largely reduce to those of a Hopfield network, and expect learning limitations to follow the existing literature on Hopfield networks \cite{hopfield_networks}. Further work will explore joint amplitude-phase encoding mechanisms, scaling to larger learned networks, and experimental validation in physical hardware.

In conclusion, this work demonstrates that signed, adaptive couplings realized through memristive synapses enable self-organized learning in ONNs without external weight-update rules or global coordination. In particular, memory formation and recall both emerge from the local coevolution of phase and conductance. This positions excitatory-inhibitory oscillator networks as a promising candidate architecture for neuromorphic systems in which learning and inference arise continuously from intrinsic device dynamics.

\section*{Acknowledgments}

We thank Siddharth Mansingh for helpful discussions and Aiping Chen for experimental support at the Center for Integrated Nanotechnologies (CINT).  
AD and FB also gratefully acknowledge support from the Center for Nonlinear Studies (CNLS) at LANL.
This work was performed under the auspices of the U.S. Department of Energy (DOE) at Los Alamos National Laboratory, operated by Triad National Security, LLC (contract 89233218CNA000001). It was supported by the DOE Advanced Scientific Computing Research (ASCR) program Award No. DE-SCL0000118. 

\bibliographystyle{unsrt}  

\bibliography{references2}

\begin{thebibliography}{10}

\bibitem{Indiveri_IEEE_2015}
Giacomo Indiveri and Shih-Chii Liu.
\newblock Memory and information processing in neuromorphic systems.
\newblock {\em Proceedings of the IEEE}, 103(8):1379--1397, 2015.

\bibitem{Kudithipudi2025}
Dhireesha Kudithipudi, Catherine Schuman, Craig~M. Vineyard, Tej Pandit, Cory Merkel, Rajkumar Kubendran, James~B. Aimone, Garrick Orchard, Christian Mayr, Ryad Benosman, Joe Hays, Cliff Young, Chiara Bartolozzi, Amitava Majumdar, Suma~George Cardwell, Melika Payvand, Sonia Buckley, Shruti Kulkarni, Hector~A. Gonzalez, Gert Cauwenberghs, Chetan~Singh Thakur, Anand Subramoney, and Steve Furber.
\newblock Neuromorphic computing at scale.
\newblock {\em Nature}, 637(8047):801--812, Jan 2025.

\bibitem{hebb_book}
D.~O. Hebb.
\newblock {\em The organization of behavior; a neuropsychological theory}.
\newblock Wiley, 1949.

\bibitem{hopfield_networks}
J.~J. Hopfield.
\newblock Neural networks and physical systems with emergent collective computational abilities.
\newblock {\em Proceedings of the National Academy of Sciences of the United States of America}, 1982.

\bibitem{exploring_neuro_computing}
Nitin Rathi, Indranil Chakraborty, Adarsh Kosta, Abhronil Sengupta, Aayush Ankit, Priyadarshini Panda, and Kaushik Roy.
\newblock Exploring neuromorphic computing based on spiking neural networks: Algorithms to hardware.
\newblock {\em ACM Comput. Surv.}, 55(12), March 2023.

\bibitem{population_doctrine}
R.~B. Ebitz and B.~Y. Hayden.
\newblock The population doctrine in cognitive neuroscience.
\newblock {\em Neuron}, 2021.

\bibitem{neuronal_oscillations}
György Buzsáki and Andreas Draguhn.
\newblock Neuronal oscillations in cortical networks.
\newblock {\em Science}, 304(5679):1926--1929, 2004.

\bibitem{patterns_oscillatory_amplitudes}
Duho Sihn and Sung-Phil Kim.
\newblock Brain-wide patterns of oscillatory amplitudes represent naturalistic behavior.
\newblock {\em NeuroImage}, 2025.

\bibitem{theta_phase_procession}
William~E. Skaggs, Bruce~L. McNaughton, Matthew~A. Wilson, and Carol~A. Barnes.
\newblock Theta phase precession in hippocampal neuronal populations and the compression of temporal sequences.
\newblock {\em Hippocampus}, 6(2):149--172, 1996.

\bibitem{theta_rhythm}
M.~E. Hasselmo and C.~E. Stern.
\newblock Theta rhythm and the encoding and retrieval of space and time.
\newblock {\em NeuroImage}, 2014.

\bibitem{what_is_a_moment}
J.~J. Hopfield and Carlos~D. Brody.
\newblock What is a moment? transient synchrony as a collective mechanism for spatiotemporal integration.
\newblock {\em Proceedings of the National Academy of Sciences}, 98(3):1282--1287, 2001.

\bibitem{attractor_dynamics_neocortex}
R.~Cossart, D.~Aronov, and R.~Yuste.
\newblock Attractor dynamics of network up states in the neocortex.
\newblock {\em Nature}, 2003.

\bibitem{memristor_chua}
Leon~O. Chua.
\newblock Memristor-the missing circuit element.
\newblock {\em IEEE Transactions on Circuit Theory}, 1971.

\bibitem{oscillatory_neurocomputers}
Frank~C. Hoppensteadt and Eugene~M. Izhikevich.
\newblock Oscillatory neurocomputers with dynamic connectivity.
\newblock {\em Phys. Rev. Lett.}, 82:2983--2986, Apr 1999.

\bibitem{Aoyagi}
Toshio Aoyagi.
\newblock Network of neural oscillators for retrieving phase information.
\newblock {\em Physical review letters}, 74:4075--4078, 06 1995.

\bibitem{NISHIKAWA2004134}
Takashi Nishikawa, Frank~C. Hoppensteadt, and Ying-Cheng Lai.
\newblock Oscillatory associative memory network with perfect retrieval.
\newblock {\em Physica D: Nonlinear Phenomena}, 197(1):134--148, 2004.

\bibitem{PhysRevE.84.046202}
Hyunsuk Hong and Steven~H. Strogatz.
\newblock Conformists and contrarians in a kuramoto model with identical natural frequencies.
\newblock {\em Phys. Rev. E}, 84:046202, Oct 2011.

\bibitem{PhysRevLett.106.054102}
Hyunsuk Hong and Steven~H. Strogatz.
\newblock Kuramoto model of coupled oscillators with positive and negative coupling parameters: An example of conformist and contrarian oscillators.
\newblock {\em Phys. Rev. Lett.}, 106:054102, Feb 2011.

\bibitem{shamsi2021hardware}
J.~Shamsi, M.~J. Avedillo, B.~Linares-Barranco, and T.~Serrano-Gotarredona.
\newblock Hardware implementation of differential oscillatory neural networks using vo2-based oscillators and memristor-bridge circuits.
\newblock {\em Frontiers in Neuroscience}, 15:674567, 2021.

\bibitem{emergence_analysis_ks_wien_bridge}
L.~Q. English, David Mertens, Saidou Abdoulkary, C.~B. Fritz, K.~Skowronski, and P.~G. Kevrekidis.
\newblock Emergence and analysis of kuramoto-sakaguchi-like models as an effective description for the dynamics of coupled wien-bridge oscillators.
\newblock {\em Phys. Rev. E}, 94:062212, Dec 2016.

\bibitem{kuramoto_model}
Juan~A. Acebr\'on, L.~L. Bonilla, Conrad~J. P\'erez~Vicente, F\'elix Ritort, and Renato Spigler.
\newblock The kuramoto model: A simple paradigm for synchronization phenomena.
\newblock {\em Rev. Mod. Phys.}, 77:137--185, Apr 2005.

\bibitem{Amer:2017}
Sherif Amer, Sagarvarma Sayyaparaju, Garrett~S. Rose, Karsten Beckmann, and Nathaniel~C. Cady.
\newblock A practical hafnium-oxide memristor model suitable for circuit design and simulation.
\newblock In {\em Proceedings of {IEEE} International Symposium on Circuits and Systems {(ISCAS)}}, Baltimore, MD, USA, May 2017.

\bibitem{discrete_calculus}
Leo J.~Grady and Jonathan R.~Polimeni.
\newblock {\em Discrete Calculus}.
\newblock Springer London, 2010.

\bibitem{negative_weights}
Dmitry Krotov and John~J. Hopfield.
\newblock Unsupervised learning by competing hidden units.
\newblock {\em Proceedings of the National Academy of Sciences}, 116(16):7723--7731, 2019.

\end{thebibliography}

\end{document}